\documentclass[10pt,twocolumn,letterpaper]{article}

\usepackage{iccv}              
\usepackage{graphicx}
\usepackage{amsmath}
\usepackage{amssymb}
\usepackage{pifont}
\usepackage{booktabs}
\usepackage{float}
\usepackage{algorithm}
\usepackage{algpseudocode}

\usepackage{multicol} 
\usepackage{adjustbox} 
\usepackage{multirow}
\usepackage{makecell}

\newcommand{\AlgoName}{Shallow-$\pi$}

\usepackage[table]{xcolor}
\definecolor{lightgray}{gray}{0.95}

\usepackage{amsmath, amssymb}
\newcommand{\xmark}{\ding{55}}
\usepackage{pifont}

\usepackage{amsmath}

\newcommand{\best}[1]{\textbf{#1}}
\newcommand{\second}[1]{\underline{#1}}

\definecolor{iccvblue}{rgb}{0.21,0.49,0.74}
\usepackage[pagebackref,breaklinks,colorlinks,allcolors=iccvblue]{hyperref}

\def\paperID{8613} 
\def\confName{ICCV}
\def\confYear{2025}
 
\title{\AlgoName: Knowledge Distillation for Flow-based VLAs
}

\author{
Boseong Jeon, Yunho Choi, Taehan Kim \\
Samsung Research\\
South Korea \\
{\tt\small junbs95@gmail.com, yunho10.choi@samsung.com, taehan11.kim@samsung.com}
}

\begin{document}  

\maketitle

\begin{abstract}

The growing demand for real-time robotic deployment necessitates {fast and on-device inference} for vision--language--action (VLA) models. 
Within the VLA literature, efficiency has been extensively studied at the token level, such as visual token pruning.
In contrast, {systematic transformer layer reduction has received limited attention}, and to the best of our knowledge, has not been explored for flow-based VLA models under knowledge distillation. 
In this work, we propose {Shallow-$\pi$}, a principled {knowledge distillation framework} that aggressively reduces the transformer depth of {both the VLM backbone and the flow-based action head}, compressing the model from $18 \rightarrow 6$ layers.
Shallow-$\pi$ achieves {over $2\times$ faster inference} with {less than $1\%$ absolute drop in success rate} on standard manipulation benchmarks, establishing {state-of-the-art performance among reduced VLA models}.
Crucially, we validate our approach through industrial-scale real-world experiments on Jetson Orin and Jetson Thor across multiple robot platforms, including humanoid systems, in complex and dynamic manipulation scenarios.
The project page can be found at \url{https://icsl-jeon.github.io/shallow-pi/}.
\end{abstract}


\begin{figure*}[t]
  \centering
  \includegraphics[width=0.9\linewidth]{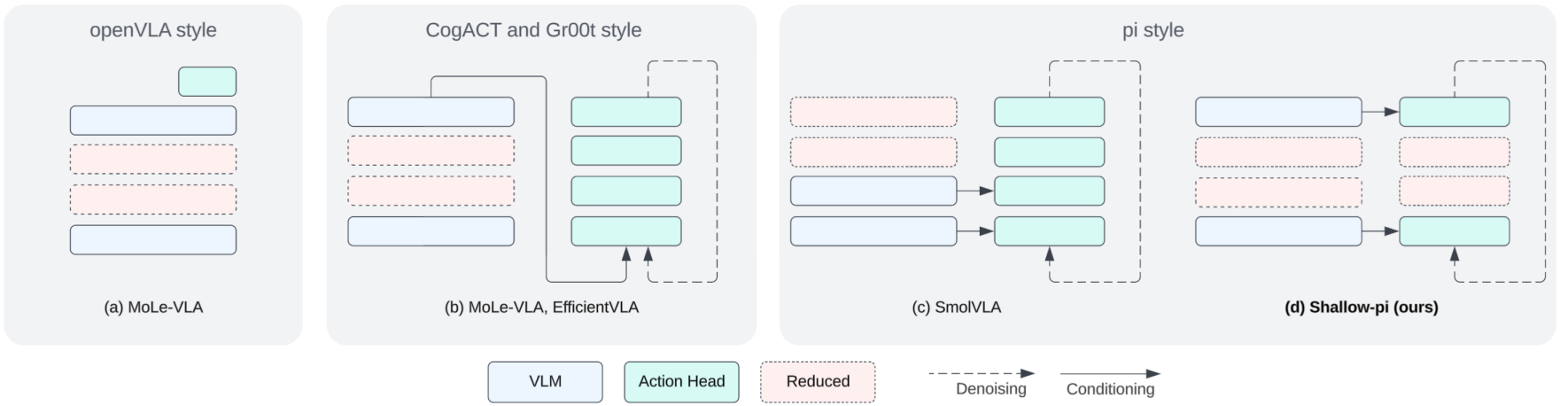}
  \caption{
Layer reduction strategies and targeted structures in the VLA domain.
Previous methods primarily reduce backbone depth only or dynamically skip layers at inference time.
In contrast, we propose a systematic knowledge distillation framework that jointly reduces the transformer depth of both the VLM backbone and the action head, which is especially effective for 
$\pi$-like architectures where the action head mirrors the VLM backbone  to receive conditioning information from all layers.
   }
  \label{fig:scope}
\end{figure*}

\section{INTRODUCTION}
Spurred by the rapid progress of large multimodal models \cite{beyer2024paligemma, chen2024internvl}, Vision-Language-Action (VLA) models \cite{bjorck2025gr00t, kim2025openvlaoft, pi0, pi05} have advanced generalist robotics by enabling diverse capabilities through a robot foundation model in an end-to-end manner. 
In these models, the vision-language model (VLM) backbone processes both images and prompts, and the resulting features are injected into the action-generation module as conditioning signals \cite{wang2025vla-adapter}.
In particular, flow-based VLA models, including $\pi$ \cite{pi0, pi05}, GR00T \cite{bjorck2025gr00t}, and CogACT \cite{li2024cogact}, adopt diffusion transformers (DiT) \cite{peebles2023scalable} as their action heads to harness the strong generative capacity of flow-matching \cite{lipman2022flow} while supporting diffusion-guidance techniques \cite{amin2025pi, park2025acg}.

However, these advantages come at a non-trivial computational cost. Flow-based VLAs combine a large VLM backbone with a diffusion-based action head consisting of dozens of transformer layers and additionally require iterative diffusion steps at inference time, which together make real-time deployment on edge devices challenging.
To mitigate these costs, prior work
has explored improving VLA efficiency along multiple axes, including reducing visual tokens \cite{vlacache, wang2025specprune, li2025cogvla}, diffusion steps \cite{yan2025maniflow, prasad2024consistency}, and transformer layers or attention computation \cite{yue2024deer, zhang2025mole, wen2025tinyvla}, as well as applying quantization \cite{park2024quantization} and graph-level optimization \cite{ma2025running}.
Among existing efficiency approaches, we focus on reducing the transformer depth of \textit{$\pi$-like} flow-based VLA architectures, which inject vision–language features into the action head at every layer \cite{wang2025vla-adapter}.

To reduce transformer computation of VLAs, in general, prior works have primarily followed two directions: 
(1) {layer-skipping methods} \cite{yue2024deer, yang2025efficientvla, zhang2025mole}, and 
(2) {using smaller backbone models} \cite{wen2025tinyvla, shukor2025smolvla}.
Layer-skipping approaches leverage inter-layer feature similarity to bypass redundant computation \cite{yue2024deer, yang2025efficientvla}, or employ router-based mechanisms that conditionally skip transformer blocks \cite{zhang2025mole}. 
However, these methods typically require the {full model to remain resident in GPU memory at inference time}, since layers are dynamically skipped rather than structurally removed.
Also, existing layer-skipping methods are generally evaluated only on backbone depth reduction, as illustrated in Figure~\ref{fig:scope}-(a,b).
This limits their applicability to \(\pi\)-like flow-based VLA models \cite{amin2025pi, du2025himoe}, where the action head commonly mirrors the VLM depth in order to consume features from intermediate layers (see Figure~\ref{fig:scope}-(c, d)).
In addition, prior evaluations are largely conducted in simulation environments \cite{yue2024deer, yang2025efficientvla} or relatively simple single-arm tasks with limited dexterity in static scenes \cite{shukor2025smolvla}, and predominantly target server-class GPUs (e.g., RTX 4090) \cite{zhang2025mole}. 

The second line of work focuses on adopting smaller VLM backbones by reducing hidden dimensions or performing early exits from intermediate layers \cite{wen2025tinyvla, shukor2025smolvla} for VLM backbone (see Figure~\ref{fig:scope}-(c)). 
While effective in lowering computational cost, these approaches often require training the VLM backbone from scratch \cite{wen2025tinyvla}, which limits their compatibility with large pretrained models and hinders performance scaling in complex manipulation tasks. 
Moreover, these methods do not reduce the depth of the action head, which is particularly critical for flow-based models where inference involves repeated denoising steps, making action-head computation a dominant cost.


In this work, we propose \textbf{\AlgoName{}} with the following contributions:
\begin{itemize}
    \item We develop and validate a knowledge distillation framework that jointly compresses {both the VLM backbone and the action head} in $\pi$-like flow-based VLAs, achieving up to {70\% layer reduction} while preserving the layer-wise feature transfer required by their architecture. 
    
    \item We carefully design and systematically ablate a set of distillation objectives—including ground-truth supervision, teacher trajectory imitation, and intermediate attention transfer—tailored to $\pi$-like flow-based VLAs, where only action tokens are denoised and multimodal features are injected from the VLM backbone at every layer.

    \item We demonstrate strong real-world performance and computational efficiency by deploying a {6-layer Shallow-$\pi$} model on edge devices across complex and dynamic manipulation tasks, achieving almost {10 Hz end-to-end inference} on Jetson Orin, without relying on graph-level optimizations or runtime conversion techniques.
\end{itemize}

\section{RELATED WORKS}

\subsection{Efficiency Challenges in VLA Models}
Prior work has explored improving the computational efficiency of Vision-Language-Action (VLA) models across several dimensions, such as fewer visual tokens \cite{jiang2025better, li2025cogvla}, fewer diffusion steps \cite{yan2025maniflow}, and reduced transformer depth or attention computation \cite{yue2024deer, zhang2025mole, wen2025tinyvla}. These directions are largely orthogonal and can be combined in principle.
A particularly active line of work \cite{vlacache, wang2025specprune, li2025cogvla, jiang2025better} focuses on reducing the number of visual tokens through pruning or caching, which can substantially reduce floating-point operations. However, the resulting wall-clock latency improvements can be workload-dependent, as modern accelerators efficiently parallelize token-wise computation \cite{fernandez2023framework, yang2025efficientvla}. 
In contrast, transformer layers are executed sequentially, and in diffusion-based VLAs the action head is invoked repeatedly across denoising steps, making model depth a critical factor for real-time inference. We defer a quantitative comparison of these two efficiency axes to Section~\ref{subsec:Impact of Layer Reduction on Latency}.

\subsection{Layer Reduction in VLA Models}

Several methods directly target transformer depth using training-free or test-time strategies.
DeeR-VLA \cite{yue2024deer} proposes an early-exit mechanism by dynamically measuring inter-layer feature similarity, while EfficientVLA \cite{yang2025efficientvla} ranks transformer layers based on cosine similarity to identify redundant computation.
However, such similarity-based criteria fail to capture the semantic role of layer depth \cite{feng2025align} (e.g., early, middle, or late layers). Moreover, existing evaluations primarily focus on decoder-only transformer architectures (e.g., OpenVLA \cite{kim2025openvla}) or reduce only the VLM backbone, even when flow-based models are considered (see Figure~\ref{fig:scope}-(a,b)).
Moreover, as observed in the image domain \cite{shen2025lazydit, you2025layer}, layer redundancy in flow-based models varies with the {noise level across diffusion steps}. 
As a result, a fixed or manually tuned threshold cannot reliably capture layer importance across both transformer depth and diffusion time.
Finally, these approaches have not been validated in real-world robotic deployments, and are typically evaluated only in simulations.

Beyond test-time thresholding, MoLE-VLA \cite{zhang2025mole} introduces a {training-based router mechanism} to conditionally skip layers. 
However, this approach is applied only to the VLM backbone and still requires the full model to remain resident in memory at inference time. 
Such conditional execution also hinders efficient batch inference, as different inputs may activate different execution paths, and typically relies on surrogate gradient estimators to train discrete routing decisions, which can introduce additional optimization instability \cite{han2021dynamic}.
Furthermore, input-dependent routing introduces {dynamic control flow}, which complicates ahead-of-time compilation and graph-level optimization on embedded platforms \cite{han2021dynamic}. 

\subsection{Knowledge Distillation in Other Domains}
Knowledge distillation has been extensively studied as an effective approach for reducing model size and inference cost while preserving performance in language, vision, and multimodal models.
In language and vision-language models, distillation methods such as DistilBERT \cite{sanh2019distilbert} and MobileVLM \cite{chu2023mobilevlm} have consistently demonstrated that compact student models distilled from large teachers outperform models trained from scratch with comparable capacity. 
More recent studies extend distillation to multimodal settings by transferring cross-modal alignment and attention structure \cite{feng2025align, kim2025compodistill}.
Distillation has also been explored for improving the efficiency of diffusion transformers by reducing model depth or structural complexity. Prior work proposes training shallow diffusion transformers \cite{shen2025lazydit, fang2025tinyfusion}, distilling contiguous groups of layers \cite{ma2025pluggable}, or adaptively selecting layers during training \cite{you2025layer}. 
However, these approaches have not been studied in VLA architectures that employ a separate vision–language backbone and a diffusion-based action head, where only action tokens participate in the denoising process. Moreover, their effectiveness has not been validated on real-world robotic systems.

\section{Preliminaries}
\subsection{Flow-based VLAs}
\label{subsec:flow_matching_vla}

Flow-based Vision--Language--Action (VLA) models generate continuous robot actions by learning a conditional vector field that transports noise to action trajectories. Given an observation $o$, language instruction $l$, and a ground-truth action $a$ 
(represented as a concatenated action chunk), a noisy interpolation is constructed as
$a_{\tau} = \tau a + (1-\tau)\epsilon$, where $\epsilon \sim \mathcal{N}(0, I)$ and 
$\tau \in [0,1]$.
 The model $v_{\theta}(a_{\tau}, o, l, \tau)$ is trained via flow matching to predict the target velocity $u = a - \epsilon$ by minimizing $\mathbb{E}[\|v_{\theta}(a_{\tau}, o, l, \tau) - u \|_2^2]$. 
Following $\pi_{0}$ \cite{pi0} and $\pi_{0.5}$ \cite{pi05}, $v_{\theta}$ consists of a vision--language transformer backbone that encodes multimodal tokens from $o$ and $l$, and a transformer-based action head that processes $a_{\tau}$ together with state information. During inference, key–value pairs from the vision–language backbone can be cached and reused across diffusion steps, since attention from vision–language tokens to action tokens is masked out, while action tokens attend to fixed vision–language representations. Actions are generated by numerically integrating the learned vector field from $\tau=0$ to $\tau=1$ using a finite number of diffusion steps.
Throughout the remainder of this paper, we adopt the notation introduced in this section when discussing flow-based VLA models.

\subsection{Impact of Layer Reduction on Latency}

\begin{figure}[t]
  \centering
  \includegraphics[width=0.998\linewidth]{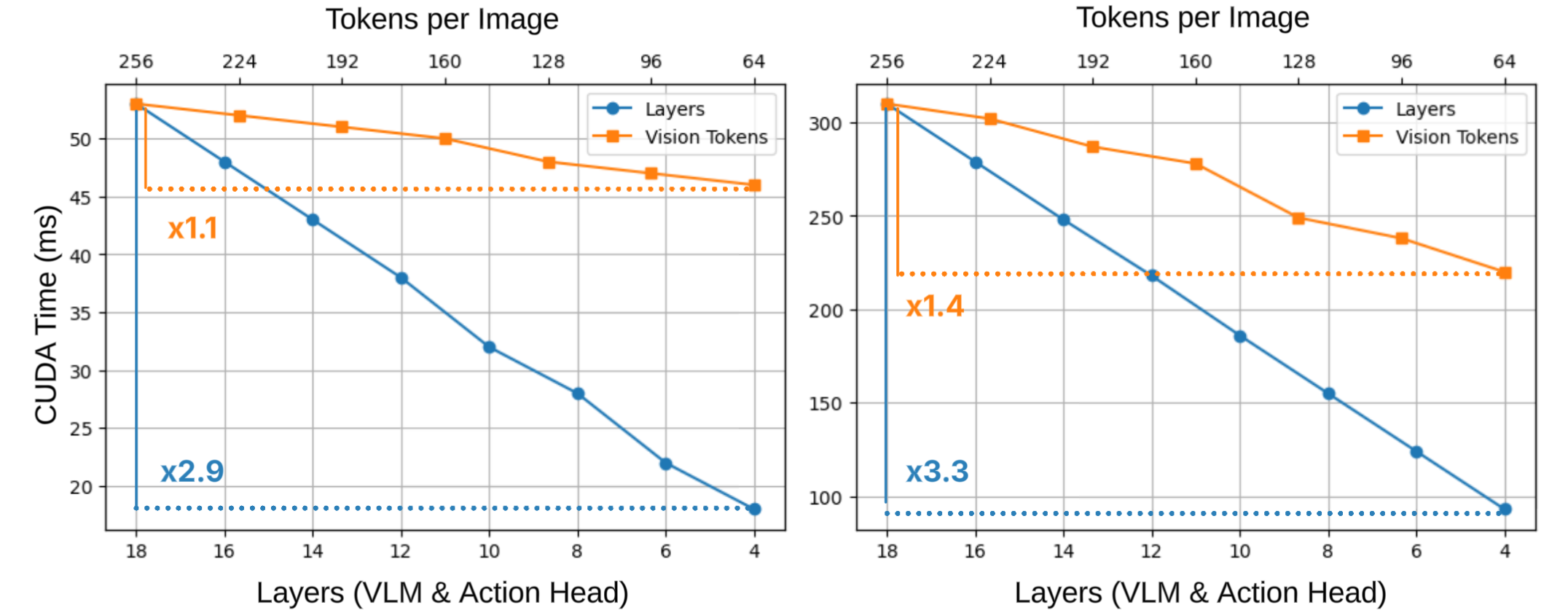}
  \caption{CUDA inference time as a function of transformer depth and visual token count. Measurements are obtained using the $\pi_{0.5}$ model trained on LIBERO, evaluated on an H100 GPU (left) and Jetson Orin (right).}
  \label{fig:latency_scaling}
\end{figure}

\label{subsec:Impact of Layer Reduction on Latency}
In this section, we analyze inference latency in flow-based VLAs by varying transformer depth and visual token count, highlighting the relative advantage of layer reduction on modern GPUs, where token-level computation is highly parallelized but transformer layers remain sequential.
Figure~\ref{fig:latency_scaling} compares the latency reduction achieved by decreasing transformer depth versus reducing the number of visual tokens per image.
For this benchmark, we use \texttt{torch.compile} to optimize inference \cite{dao2022flashattentionfastmemoryefficientexact}, and report results on both H100 and Jetson Orin, representing the high and low ends of computational capability.
We set the maximum reduction boundaries to 4 transformer layers and 64 visual tokens, following the most aggressive reductions reported in recent work \cite{chu2023mobilevlm, yang2025efficientvla}.

While token reduction yields only modest latency improvements, reducing the number of transformer layers leads to a substantially larger decrease in inference time.
This trend reflects the highly parallel nature of token-level computation on modern GPUs, in contrast to the strictly sequential execution of transformer layers, whose costs accumulate directly in wall-clock time.
This effect is particularly evident on high-performance hardware such as the H100, where token reduction provides minimal benefit.
Overall, these results highlight the unique effectiveness of transformer layer reduction as a means of improving inference latency.

\begin{figure}[t]
  \centering
    \includegraphics[width=0.96\linewidth]{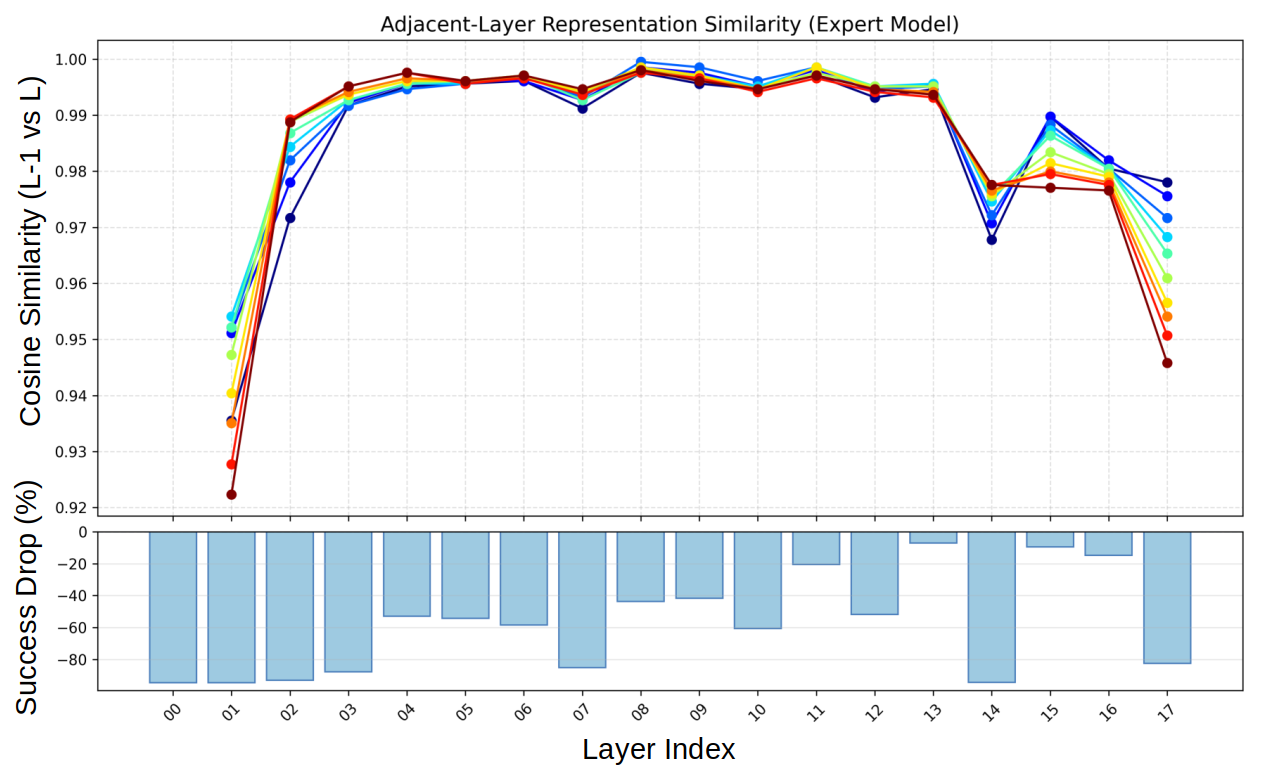}
  \caption{(Top) Feature similarity trend along noise level $\tau$. (Bottom) Layer sensitivity analysis of $\pi_{0.5}$ on the LIBERO benchmark. The bar chart shows the decrease in average success rate caused by skipping individual layers.}
  \label{fig:layer-analysis}
\end{figure}
\section{Why Layer Skipping Is Insufficient}
Motivated by the computational advantages of transformer depth reduction,
 prior work proposes \emph{test-time layer skipping} for OpenVLA-style models based on feature similarity or learned routing mechanisms \cite{yue2024deer, zhang2025mole, yang2025efficientvla, yangdysl}, either skipping redundant layers or dynamically selecting layers to bypass.
We examine whether these approaches remain effective for $\pi$-like flow-based VLAs, which inject visual–language features at every transformer layer.
As a representative example of skipping based on feature similarity, EfficientVLA \cite{yang2025efficientvla} uses cosine similarity between adjacent layers to decide whether layers can be skipped. As depicted in Figure \ref{fig:layer-analysis}, we analyze layer-wise cosine similarity of $\pi_{0.5}$ across denoising timesteps $\tau$ and perform layer sensitivity analysis measuring averaged LIBERO success rate when individual layers are skipped. 

Our results reveal two limitations of this approach. First, similarity profiles vary substantially with $\tau$, making fixed skipping rules brittle. For example, although overall similarity at $\tau=0$ is lower than at $\tau=1$ in early and late layers, this trend is not consistent across all depths (e.g., layer $13 \rightarrow 14$), preventing a simple or monotonic thresholding strategy conditioned on $\tau$. Second, similarity poorly predicts functional importance: despite higher similarity between layers 1 and 2 than between layers 16 and 17, skipping the former leads to a much larger drop in success rate. This confirms that similarity-based criteria alone are insufficient, consistent with prior findings that different transformer layers serve distinct roles \cite{feng2025align}.

\begin{figure}[t]
  \centering
    \includegraphics[width=0.92\linewidth]{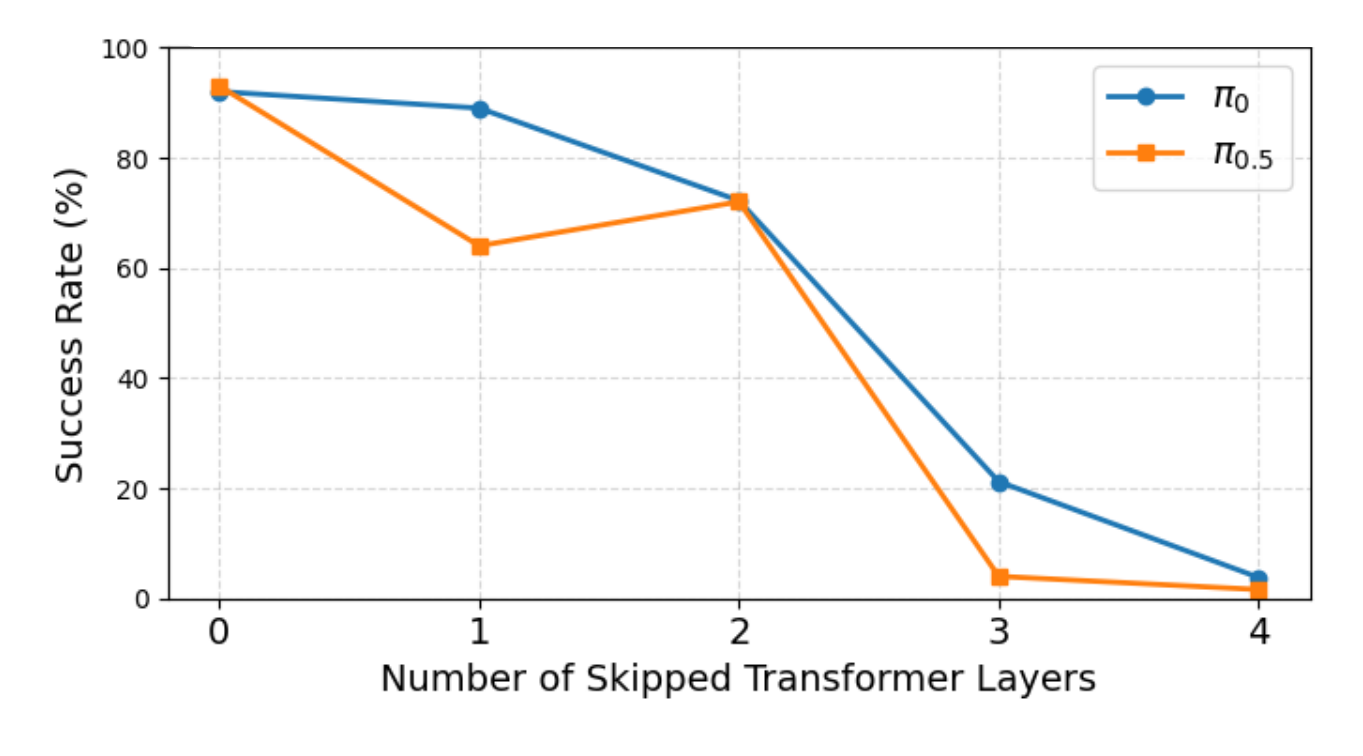}
  \caption{Success rate (\%) on LIBERO (Spatial, Object, Goal, and 10) as a function of the number of skipped transformer layers.}
  \label{tab:layer_skipping_sr}
\end{figure}

We further consider learned routing as an alternative by using layer sensitivity as an oracle for layer selection. Progressively removing layers in order of lowest sensitivity (e.g., $13 \rightarrow 15 \rightarrow 16 \rightarrow 11 \rightarrow \cdots$) causes the success rate to collapse once more than three layers are removed (Figure ~\ref{tab:layer_skipping_sr}). Taken together, these results indicate that test-time layer skipping—via similarity metrics or routing—is fundamentally limited for flow-based VLAs, where layer functionality is tightly coupled with denoising dynamics \cite{shen2025lazydit, you2025layer}. 
This motivates our use of \emph{knowledge distillation} for effective and aggressive layer reduction, rather than relying on complex threshold heuristics or routing mechanisms.

\section{Knowledge Distillation for Shallow Layers}

\begin{figure}[t]
  \centering
  \includegraphics[width=0.95\linewidth]{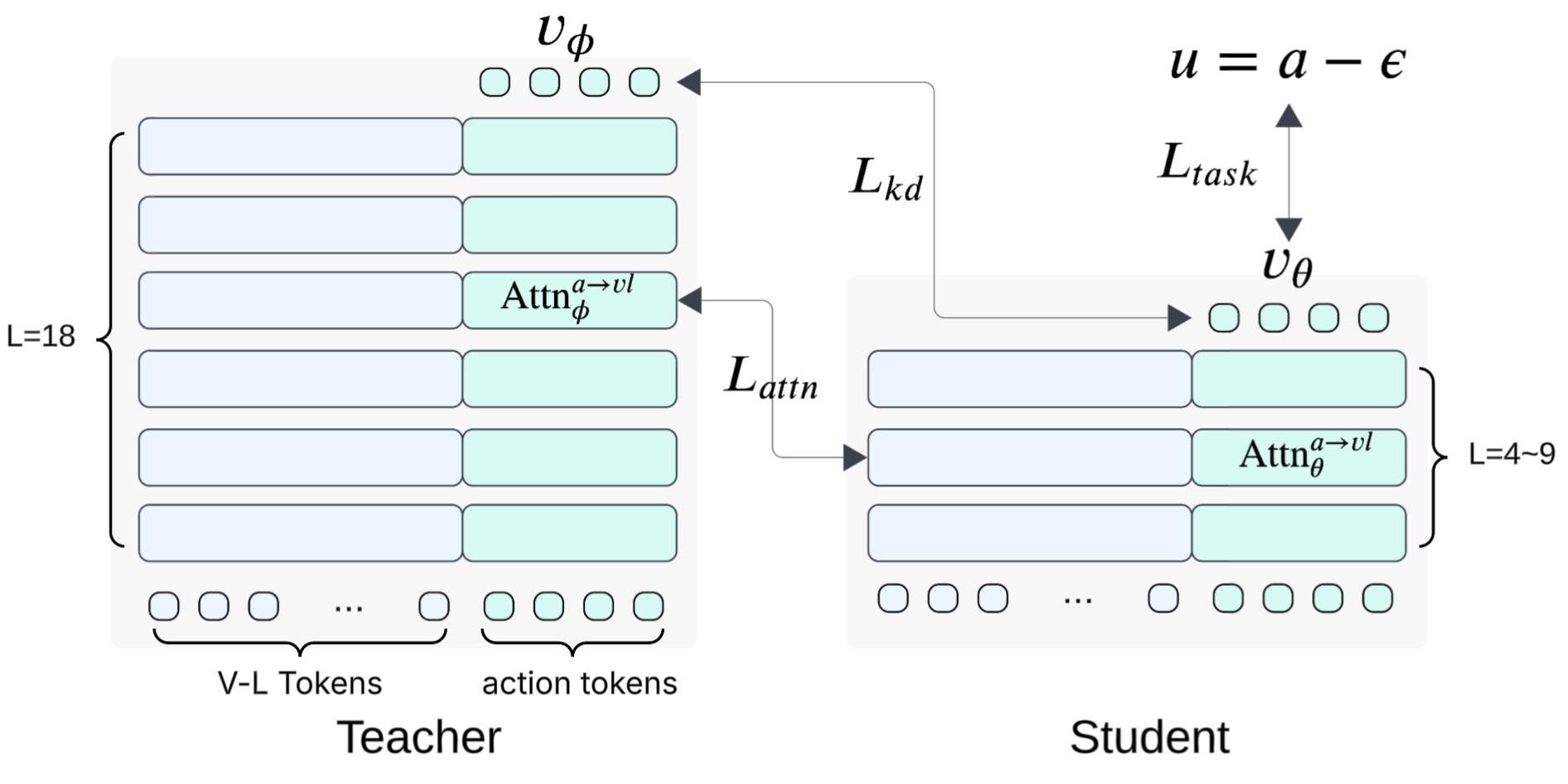}
  \caption{Shallow-$\pi$ reduces the transformer depth of the VLM backbone and action head via knowledge distillation, using three loss terms to match ground-truth actions, teacher outputs, and intermediate cross-attention between the backbone and action head.}
  \label{fig:distill_overview}
\end{figure}

Given a pretrained flow-based VLA policy $v_{\phi}$, our goal is to obtain a student model $v_{\theta}$ with substantially fewer transformer layers, while preserving action generation performance.
The student is trained to approximate the teacher’s denoising behavior through a combination of task supervision and knowledge distillation losses, as illustrated in Figure~\ref{fig:distill_overview}. 
To initialize the shallow student, we reduce the number of transformer layers in both the vision--language backbone and the action head using a uniform subsampling strategy, following TinyBERT-style layer initialization \cite{jiao2020tinybert}. We observe no additional benefit from selecting initialization layers based on the sensitivity analysis in the bottom of Figure~\ref{fig:layer-analysis}, provided the model is trained for a sufficient number of steps.

\subsection{Distillation Objectives}

We train the student model $v_{\theta}$ using a combination of three complementary losses. 
First, the task loss $L_{\text{task}}$ follows standard flow matching and supervises the student to predict the ground-truth velocity.

\begin{equation}
    L_{\text{task}} = \mathbb{E}[\|v_{\theta}(\cdot) - u \|_2^2]
\end{equation}

Second, a knowledge distillation loss $L_{\text{kd}}$ encourages the student to match the teacher’s predicted velocity, offering informative teacher-generated guidance.

\begin{equation}
    L_{\text{kd}} = \mathbb{E}[\|v_{\theta}(\cdot) - v_{\phi}(\cdot) \|_2^2]
\end{equation}

Finally, we introduce a novel attention distillation loss $L_{\text{attn}}$ that is carefully designed for the multimodal VLA architecture, aligning the cross-attention distributions between action queries and vision--language key–value pairs at an intermediate transformer layer:

\begin{equation}
\label{eqn:L_attn}
L_{\text{attn}} =
\mathbb{E}\!\left[
\operatorname{KL}\!\left(
\operatorname{Attn}_{\phi}^{a \rightarrow  vl}
\;\|\;
\operatorname{Attn}_{\theta}^{a \rightarrow  vl}
\right)
\right],
\end{equation}
where $\operatorname{Attn}^{a \rightarrow vl} = 
\mathrm{softmax}\!\left(Q^{a} K^{vl \top}\right)$, and
the $\operatorname{KL}$ is KL divergence is evaluated over the corresponding attention distribution across vision–language tokens for each action token.
Together, these three losses promote both output-level consistency and representation alignment between the teacher and the shallow student.

\subsection{Attention Distillation}

In contrast to prior attention distillation approaches in LLMs and VLMs, we do not enforce attention matching over the full token set (i.e., $\operatorname{Attn}^{vl \rightarrow vl}$). 
This choice is motivated by the structure of flow-matching VLAs, where only action tokens define the generative policy, while visual and language tokens from the VLM backbone serve purely as conditioning context. 
Distilling attention over non-generated backbone tokens over-constrains the student and introduces interference with pre-trained representations, without improving control fidelity. 
Empirically, we observe that matching attention across all tokens consistently leads to unstable training and near-zero success rates in the final policy (see Table~\ref{tab:ablations}-(c)).
Also, for more efficient and flexible optimization, we take inspiration from Align-KD \cite{feng2025align} which applies attention transfer only at a specific layer, rather than for the whole layers. 
We apply attention distillation at a middle transformer layer rather than at early or late stages. 
Because the student is initialized by directly copying the bottom layers of the teacher, early-layer representations are already closely aligned and provide limited additional supervision. 
At the final layer, output matching is already enforced through $L_{\text{task}}$ and $L_{\text{kd}}$, making further alignment redundant.

\subsection{Ablations}
Based on the $\pi_{0.5}$ model trained on the LIBERO benchmark \cite{liu2023libero}, we conduct ablation studies on both the loss components and the transformer layer at which $L_{\text{attn}}$ in Eq.~\ref{eqn:L_attn} is applied. Across all experiments, we use the same teacher model and train the student for 30K steps with a batch size of 64. As shown in Table~\ref{tab:ablations}, the composite loss formulation combined with applying attention distillation at the middle layer yields the best performance.

\begin{table}[t]
\centering
\caption{Ablations for loss design}
\label{tab:ablations}
\small
\renewcommand{\arraystretch}{1.1}

\begin{subtable}{\linewidth}
\centering
\caption{Loss ablation on Shallow $\pi_{0.5}$ with different depths}
\begin{tabular}{lccc ccc}
\toprule
$\mathcal{L}_{\text{task}}$ 
& $\mathcal{L}_{\text{kd}}$ 
& $\mathcal{L}_{\text{attn}}$ 
& \textbf{L9} 
& \textbf{L6} 
& \textbf{L4} \\
\midrule
\checkmark & \xmark     & \xmark     & 95.4 & 93.0 & 92.3 \\
\checkmark & \checkmark& \xmark     & 96.2 & 93.9 & 93.9 \\
\rowcolor{lightgray}
\checkmark & \checkmark& \checkmark & \textbf{96.8} & \textbf{94.6} & \textbf{94.2} \\
\bottomrule
\end{tabular}
\end{subtable}
\vspace{6pt}

\begin{subtable}{\linewidth}
\centering
\caption{Attention distillation placement (L=number of layers).}
\begin{tabular}{lccc}
\toprule
 & \textbf{L9} & \textbf{L6} & \textbf{L4} \\
\midrule
Initial layer & 94.4 & 93.9 & 91.0 \\
\rowcolor{lightgray}
Middle layer  & \textbf{96.8} & \textbf{94.6} & \textbf{94.2} \\
Later layer   & 95.6 & 94.1 & 92.8 \\
\bottomrule
\end{tabular}
\end{subtable}
\vspace{6pt}

\begin{subtable}{\linewidth}
\centering
\caption{Attention distillation target queried tokens.}
\begin{tabular}{lccc}
\toprule
 & \textbf{L9} & \textbf{L6} & \textbf{L4} \\
\midrule
\rowcolor{lightgray}
Action tokens only   & {96.8} & 94.6 & 93.9 \\
All tokens    & 0.0 & 61.5 & 0.0 \\
\bottomrule
\end{tabular}
\end{subtable}

\end{table}

\begin{figure}[t]
  \centering
  \includegraphics[width=0.92\linewidth]{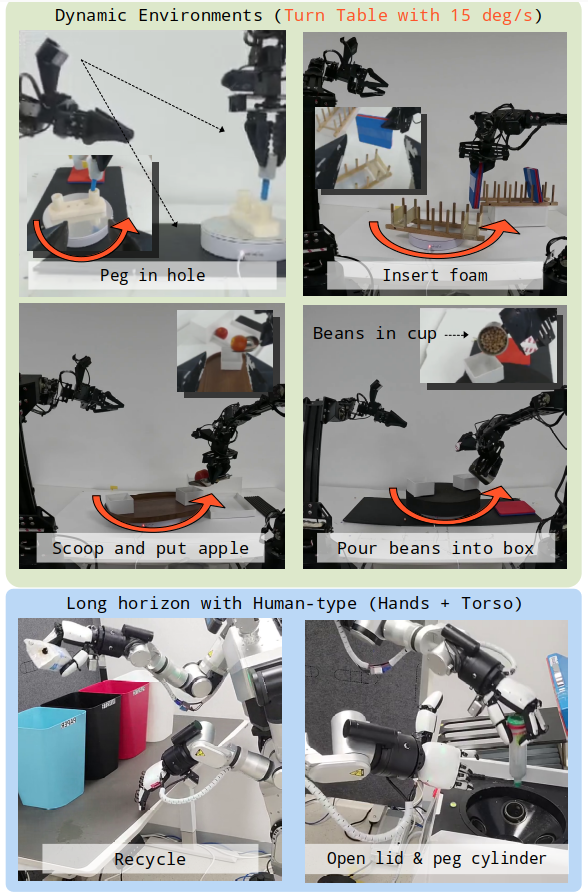}
  \caption{Experiment task suites for Shallow-$\pi$. The benchmarks include scenarios requiring complex manipulation under dynamic scenes, as well as tasks involving articulated robot platforms with coordinated hand and torso movements. All real-world evaluations are performed on edge devices (Jetson Thor and Orin).
   }
  \label{fig:experiment_overview}
\end{figure}

\section{Experiment}
We validate \AlgoName{} on both simulation benchmarks and real-world experiments across multiple challenging scenarios. Through these evaluations, we aim to answer the following questions:

\begin{itemize}
\item Despite aggressive transformer layer reduction and latency improvement, does \AlgoName{} preserve the teacher’s capabilities across a broad range of task prompts?
\item Is knowledge distillation a more effective approach than training a compact model from scratch?
\item Does \AlgoName{} demonstrate sufficient real-world performance when deployed on edge devices for complex and dynamic tasks?
\item Does \AlgoName{} retain generalization capability on unseen tasks without overfitting to the distillation dataset?
\end{itemize}

\subsection{Simulation Benchmark}
We evaluate our method on the LIBERO benchmark using simulation. Teacher models based on $\pi_{0}$ and $\pi_{0.5}$ are trained for 30k steps with a batch size of 64, using only two input images (a third-person camera and a wrist camera). 
We changed the official code implementation of \cite{pi0, pi05} to skip the third image input entirely, rather than leaving it as a black image. 
The student models are distilled using the same training setup. The results are summarized in Table~\ref{tab:vla_comparison}.

As shown in the table, the distilled models achieve success rates within 1\% drop of their teachers, while reducing both FLOPs and CUDA inference time by more than a factor of two. Although orthogonal to our direction, shallow flow models require significantly less computation—in terms of both FLOPs and CUDA latency—than state-of-the-art token compression methods for comparable success rates. 
Distilling high-capacity flow models also yields marginally better performance than small-backbone approaches such as SmolVLA~\cite{shukor2025smolvla}. In addition, shallow-$\pi$ models exhibit lower inference latency than SmolVLA, whose action head contains 16 transformer layers.
These results confirm that distilling high-capacity models is more effective than training small backbones from scratch, and that jointly reducing the depth of both the VLM and action head leads to superior efficiency. 
Overall, considering its straightforward implementation—without complex manual layer selection~\cite{yang2025efficientvla,yue2024deer} or sophisticated routing mechanisms~\cite{zhang2025mole,li2025cogvla}—shallow-$\pi$ achieves the best trade-off between success rate and computational efficiency.

\begin{figure}[t]
  \centering
  \includegraphics[width=0.85\linewidth]{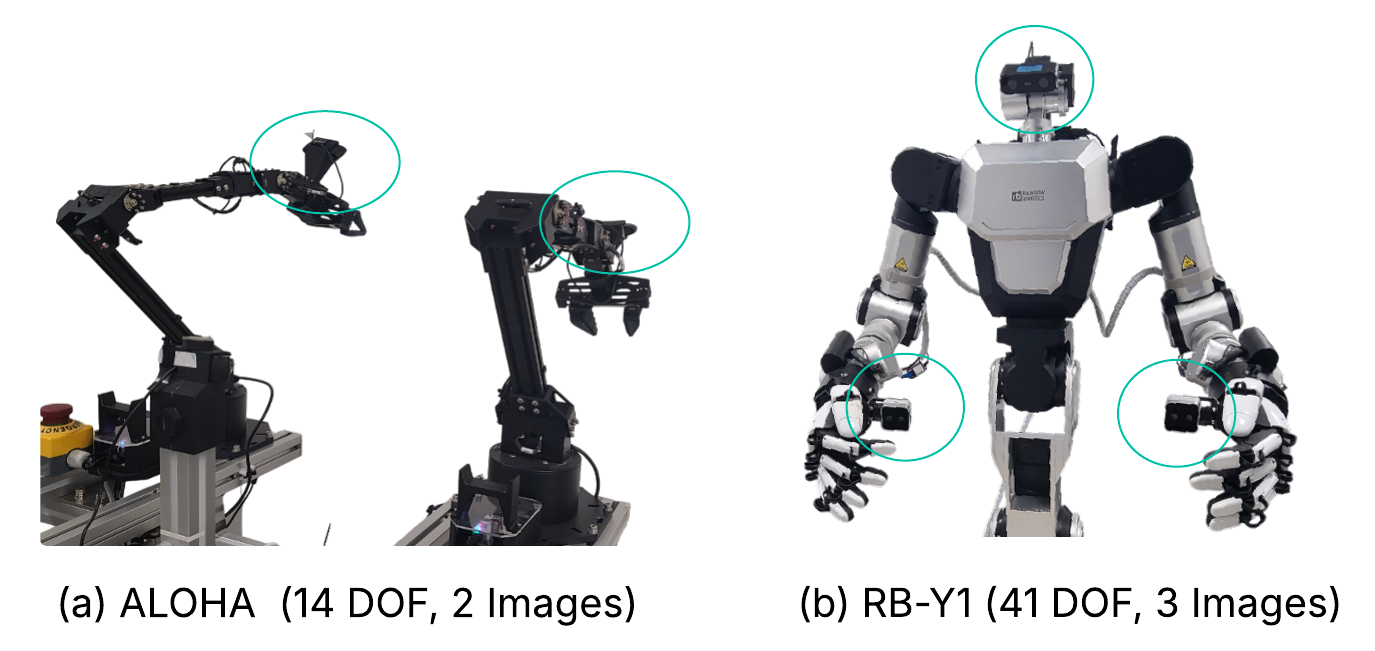}
\caption{Degrees of freedom (DoFs) and camera configurations (green circles) for the robot platforms.}
  \label{fig:robots}
\end{figure}

\begin{figure}[t]
  \centering
  \includegraphics[width=0.99\linewidth]{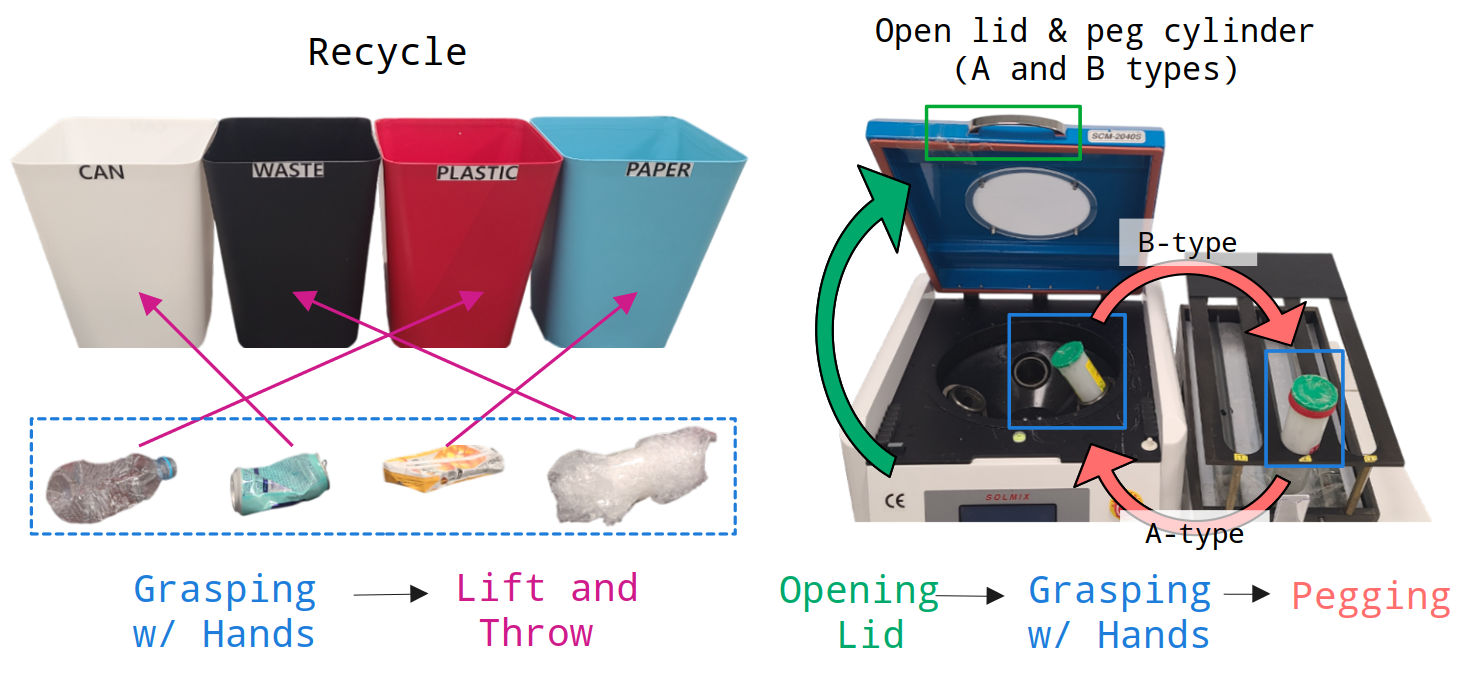}
\caption{Skill sequences required by the tasks of the hand-typed robot (RB-Y1).}
\label{fig:rby1-sequence}

\end{figure}

\begin{table*}[t]
\centering
\caption{Success rates (SR) are reported on the LIBERO benchmark, and computation is measured on an NVIDIA H100. 
Best results are shown in bold, and second-best results are underlined. All numbers are rounded.}
\label{tab:vla_comparison}

\small
\setlength{\tabcolsep}{6pt}
\renewcommand{\arraystretch}{1.05}
\begin{tabular}{llccccccc}
\toprule
\textbf{Group} & \textbf{Model} 
& \textbf{Spatial} & \textbf{Object} & \textbf{Goal} & \textbf{Long (10)} 
& \textbf{Avg} $\uparrow$
& \textbf{FLOPs (T)}$\downarrow$
& \textbf{CUDA Time (ms)}$\downarrow$ \\
\midrule

\multirow{2}{*}{Baseline }
& $\pi_{0}$ \cite{pi0}
& 97 & 97 & 93 & 92 & 95 & 2.93 & 22.6 \\
& $\pi_{0.5}$ \cite{pi05}
& \second{98} & 96 & \second{97} & 93 & 96 & 3.39 & 25.5 \\

\midrule

\multirow{2}{*}{\makecell[l]{Token Compression\\(orthogonal to ours)}}
& CogVLA \cite{li2025cogvla}
& \best{99} & \best{99} & 97 & \best{95} & \best{97} & 2.70 & 31.0 \\
& LightVLA \cite{jiang2025better}
& \second{98} & \second{98} & \best{98} & \best{95}
& \best{97} & 2.91 & 22.0 \\

\midrule

\multirow{1}{*}{Small Backbone}
& SmolVLA \cite{shukor2025smolvla}
& 90 & 96 & 92 & 71 & 87 & \best{0.50} & 26.0 \\

\midrule

\multirow{4}{*}{\makecell[l]{Layer Distillation\\(ours)}}
& $\pi_{0}$-L9
& \second{98} & \second{98} & 95 & 87 & 95 & 1.62 & 13.5 \\
& $\pi_{0}$-L6
& \second{98} & 95 & 93 & 85 & 94 & \second{1.18} & \best{10.5} \\
\cmidrule(lr){2-9}
& $\pi_{0.5}$-L9
& \best{99} & \second{98} & \second{97} & 93 & \best{97} & 1.82 & 14.8 \\
& $\pi_{0.5}$-L6
& \second{98} & 96 & 94 & \second{90} & 95 & 1.30 & \second{11.3} \\

\bottomrule
\end{tabular}
\end{table*}

\begin{table}[t]
\centering
\caption{Task performance across dynamic, complex, and unseen settings.}
\label{tab:realworld_tasks_reflected}
\small
\setlength{\tabcolsep}{10pt}
\renewcommand{\arraystretch}{1.15}

\begin{tabular}{lccc}
\toprule
\multicolumn{4}{c}{Dynamic Tasks (Jetson Orin)} \\
\midrule
Task & SmolVLA & $\pi_0$ & Shallow-$\pi_0$ \\
\midrule
Peg in hole   & 0/10 & 7/10  & \textbf{10/10} \\
Insert foam   & 1/10 & 5/10  & \textbf{7/10}  \\
Scoop apple   & 5/10 & 6/10  & \textbf{9/10}  \\
Pour beans    & 6/10 & 5/10  & \textbf{8/10}  \\
\midrule
E2E Comp. (ms) & 230 & 364 & \textbf{110} \\
\bottomrule
\end{tabular}

\vspace{6pt}

\begin{tabular}{lcc}
\toprule
\multicolumn{3}{c}{Hands and Torso (Jetson Thor)} \\
\midrule
Task & $\pi_{0.5}$  & Shallow-$\pi_{0.5}$ \\
\midrule
Recycle             & 12/20 & \textbf{17/20} \\
Open lid \& peg (A) & 5/10  & \textbf{7/10}  \\ 
Open lid \& peg (B) & 1/5   & \textbf{5/5}   \\ 
\midrule
E2E Comp. (ms)  & 130 & \textbf{78}\\
\bottomrule
\end{tabular}

\vspace{6pt}

\begin{tabular}{lcc}
\toprule
\multicolumn{3}{c}{Unseen Environments} \\
\midrule
Task & $\pi$ & Shallow-$\pi$ (L6) \\
\midrule
Peg in hole ($\pi_0$)        & 0/5  & \textbf{3/5} \\
Recycle ($\pi_{0.5}$)       & 8/20  & \textbf{15/20} \\
\bottomrule
\end{tabular}

\end{table}

\begin{figure}[t]
  \centering
  \includegraphics[width=0.75\linewidth]{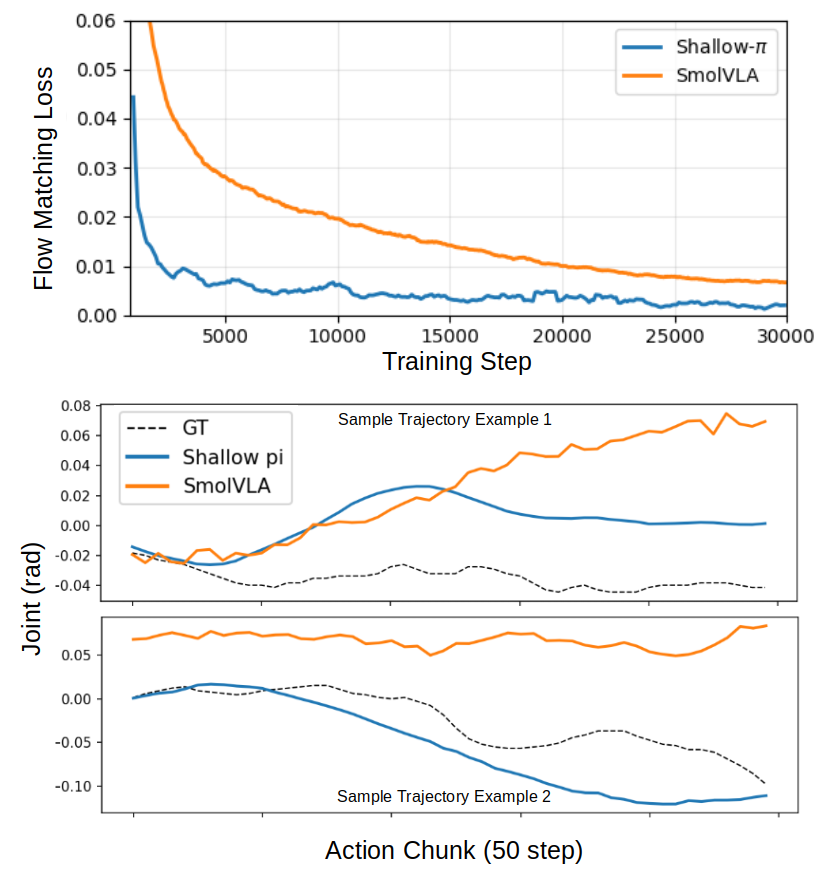}
\caption{(Top) Flow matching loss vs training step. (Bottom) Samples of chunk to compare the motion quality in the real-world training set.}
  \label{fig:jerkness}
\end{figure}

\subsection{Real-World Experiment}
We evaluate Shallow-$\pi$ in challenging scenarios requiring high computational efficiency, precise control, and strong generalization. 
As shown in Figure~\ref{fig:experiment_overview}, experiments are conducted on two robot platforms, ALOHA \cite{fu2024mobile} and RB-Y1, using only onboard sensors without static third-person cameras (see Figure~\ref{fig:robots}).
In experiments with ALOHA, we primarily rely on the left wrist camera to observe salient objects and target locations, as shown in the \textit{Peg in hole} task in Figure~\ref{fig:experiment_overview}.
We train separate models for ALOHA ($\pi_0$) and RB-Y1 ($\pi_{0.5}$) to account for their different degrees of freedom and embodiments. 
For both teacher training and distillation, we use a batch size of 128 and train for 100K steps. 
During distillation, the student model is configured with 6-transformer layers.
We perform model inference on Jetson Orin and Jetson Thor for the ALOHA and RB-Y1 platforms, respectively.
We compare Shallow-$\pi$ with its teacher and SmolVLA \cite{shukor2025smolvla}.
SmolVLA is trained until the convergence, allowing all parameters, including the vision encoder, to be trainable following the findings in \cite{kim2025openvla}.
For real-robot deployment, we use an action chunk size of 50 with a 30\,Hz control loop. 
At each step, the robot executes 7 actions before requesting the next model inference in a receding-horizon manner, continuing to execute the remaining actions from the previous chunk while awaiting the new prediction. 
Temporal ensembling \cite{zhao2023learning} is applied to ensure smooth action execution.
As noted in \cite{black2025real}, slow inference increases open-loop execution on stale observations, which can lead to non-recoverable failure states. 
High inference latency also amplifies discrepancies between old and newly predicted action chunks, reducing execution precision when temporal smoothing is applied.

As depicted in Figure \ref{fig:experiment_overview},  we design the following 6 test cases. 

\begin{itemize}
\item \textit{Peg in hole} (92 episodes): The ALOHA robot approaches and grasps a cylindrical peg, then inserts it into dynamically moving small holes. 
\item \textit{Insert foam} (100 episodes): The ALOHA robot grasps a square foam block and places it onto a moving bookshelf. The foam must be released when its edge is aligned with the shelf geometry.
\item \textit{Scoop and place apple} (190 episodes): The ALOHA robot grasps a scooping tool to collect an apple and then deposits it into a moving box.
\item \textit{Pour beans into box} (100 episodes): The ALOHA robot grasps the handle of a cup filled with beans and pours the contents into a moving box.
\item \textit{Recycle} (2{,}600 episodes total): The RB-Y1 robot grasps trash objects and throws them into the appropriate bins. 
The training set includes a diverse set of trash objects, while evaluation is conducted on a fixed subset (see the skill sequence in Figure~\ref{fig:rby1-sequence}). 

\item \textit{Open lid \& peg cylinder (Type A)}  (436 episodes): The RB-Y1 robot opens a lid, extracts a cylinder using appropriate finger configurations, and places it into a corresponding hole. 
\item \textit{Open lid \& peg cylinder (Type B)}  (214 episodes): The robot extracts a cylinder from a confined hole using torso rotation and places it into the target location. 
Compared to Type~A, this variant requires precise arm control to extract a deeply inserted cylinder.

\end{itemize}

\begin{figure}[t]
  \centering
  \includegraphics[width=0.8\linewidth]{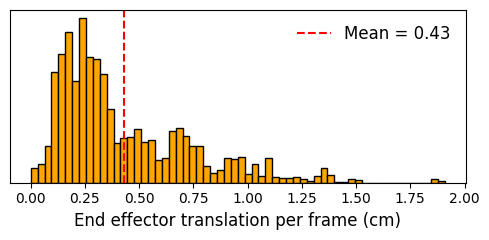}
\caption{Histogram of the per-frame average translation of the right-arm end effector in the training dataset. }
  \label{fig:openloop-histogram}
\end{figure}

\begin{figure}[t]
  \centering
  \includegraphics[width=0.95\linewidth]{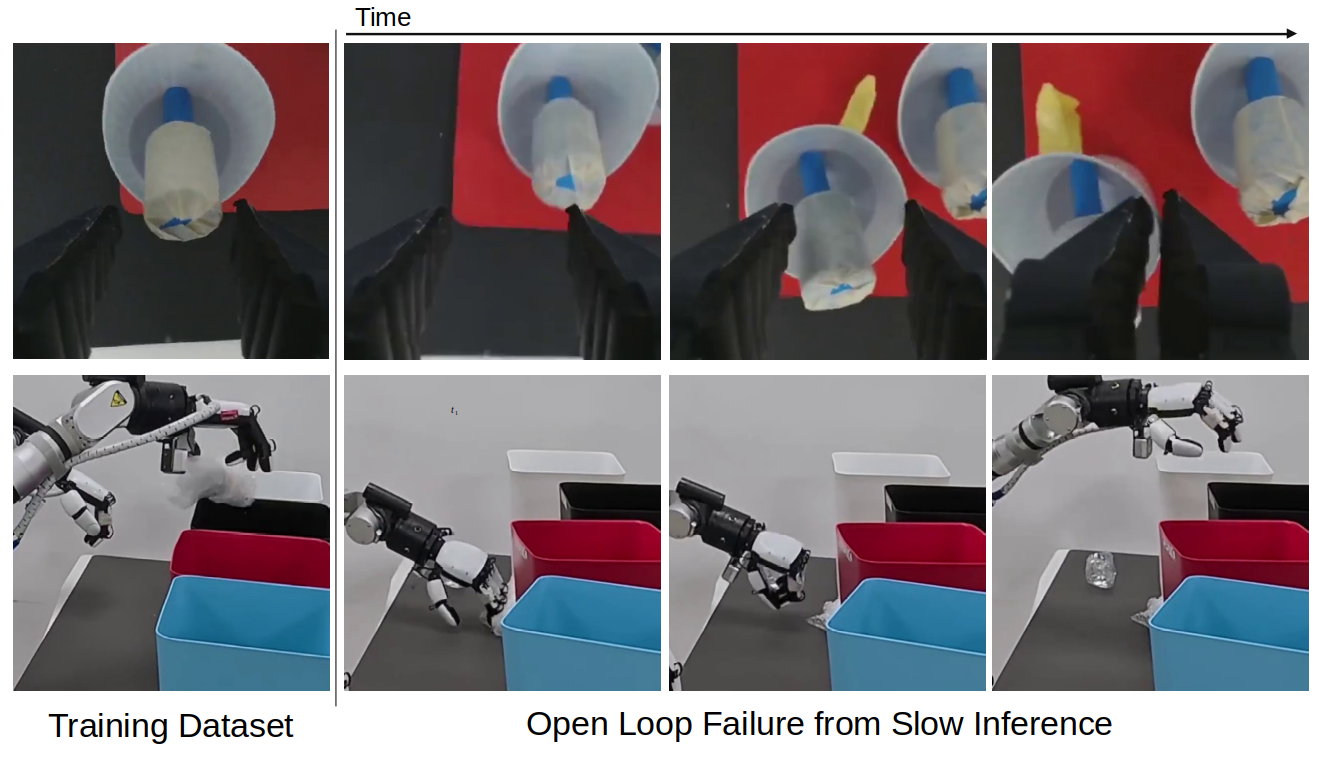}
\caption{Experimental snapshots illustrating open-loop failures of the teacher model, where the student model achieves better performance by reacting to observations more rapidly.}
  \label{fig:openloop-failure}
\end{figure}

As summarized in Table~\ref{tab:realworld_tasks_reflected}, Shallow-$\pi$ consistently achieves better real-world performance across all tasks, primarily due to reduced inference latency while preserving model capability.
For the ALOHA platform, the distilled model reduces inference time by more than 200\,ms (approximately 6 frames) compared to the teacher, which corresponds to over 2\,cm of additional open-loop translation of the end-effector on average, as illustrated in Figure~\ref{fig:openloop-histogram}.
Given that the linear speed at the edge of the turntable is approximately 0.8\,cm/s, this prolonged open-loop motion measurably degrades placement precision, a trend that is reflected in the experimental results.
As shown in the training curves in Figure~\ref{fig:jerkness}, SmolVLA exhibits a flow-matching loss approximately twice that of Shallow-$\pi$, and correspondingly produces noisier and less precise actions, as illustrated in the bottom row of Figure~\ref{fig:jerkness} and in Table~\ref{tab:realworld_tasks_reflected}
In contrast, the distilled Shallow-$\pi$ model demonstrates strong robustness and generalization, achieving over 80\% success on the recycle task, which requires reliable grasping of objects across diverse poses.

We further investigate whether the proposed distillation preserves generalization capacity without overfitting to the training data despite the reduced model size. 
To this end, we evaluate the models under two unseen perturbation scenarios, as shown in Figure \ref{fig:openloop-failure}. 
First, we shift the initial position of the cylinder in the \textit{peg in hole} task on ALOHA by 3\,cm, creating an unseen object configuration. 
Second, we displace each trash bin by more than 10\,cm in the \textit{recycle} task.
We observe that increased open-loop execution in the teacher model often prevents correction of picking or releasing poses, as illustrated in the first row of Figure~\ref{fig:openloop-failure}. 
In contrast, the distilled model achieves better performance by incorporating updated visual observations more frequently, demonstrating improved robustness and generalization under unseen spatial perturbations.

\section{CONCLUSIONS}
In this work, we introduced Shallow-$\pi$, an efficient knowledge distillation framework for flow-based VLA models that inject conditioning information at all intermediate layers. 
Our results show that the distilled model preserves the performance and generalization capability of the teacher while significantly reducing computation through aggressive transformer layer reduction. 
Also, compared to the models with a small backbone, the distilled model acheived better performance in denoising and the precision.
We validated the effectiveness of Shallow-$\pi$ through real-world deployment on edge devices, demonstrating reliable performance under practical latency constraints. 

Our work is not without limitations. First, compared to layer-skipping approaches, knowledge distillation incurs higher training-time computational costs, as both teacher and student models must be loaded simultaneously. To mitigate this overhead, future work should investigate more effective strategies for selectively freezing model components to reduce VRAM consumption during distillation, as well as curating or filtering key training samples that provide the most informative supervision. Looking ahead, we also plan to explore complementary efficiency axes—such as visual token reduction and diffusion step reduction—to further improve inference throughput.

{
    \small
    \bibliographystyle{ieeenat_fullname}
    \bibliography{main}
}


\end{document}